\documentclass[a4paper,11pt]{article}

\usepackage{a4}

\usepackage{hyperref}
\usepackage[sort&compress,numbers]{natbib}

\usepackage{amssymb}
\usepackage{amsmath}
\usepackage{tikz}
\usepackage{pgfplots}

\usepackage[ruled]{algorithm2e}
\usepackage{float}

\begin{document}

\title{Towards data-driven filters in Paraview}
\author{Drishti Maharjan\thanks{Department of Computer Science \& Electrical Engineering,
Jacobs University Bremen gGmbH, Campus Ring 1, 28759 Bremen, Germany} \and Peter Zaspel\footnotemark[1]}
\date{}

\maketitle

\begin{abstract}
Recent progress in scientific visualization has expanded the scope of visualization from
being merely a way of presentation to an analysis and discovery tool. A given visualization
result is usually generated by applying a series of transformations or filters to the
underlying data. Nowadays, such filters use deterministic algorithms to process the data.
In this work, we aim at extending this methodology towards data-driven filters, thus filters
that expose the abilities of pre-trained machine learning models to the visualization
system. The use of such data-driven filters is of particular interest in fields like
segmentation, classification, etc., where machine learning models regularly outperform
existing algorithmic approaches. To showcase this idea, we couple Paraview, the well-known
flow visualization tool, with PyTorch, a deep learning framework. Paraview is extended by
plugins that allow users to load pre-trained models of their choice in the form of newly
developed filters. The filters transform the input data by feeding it into the model and
then provide the model's output as input to the remaining visualization pipeline.
A series of simplistic use cases for segmentation and classification on image and fluid data is
presented to showcase the technical applicability of such data-driven transformations
in Paraview for future complex analysis tasks. 
\end{abstract}


\section{Introduction}

 \label{Introduction}

      
        Over the recent years, \textit{machine learning} has emerged to an important tool for data analysis and prediction. Due to the tremendous success of many machine learning algorithms, machine learning is being used in almost every field ranging from industry \cite{industry} to health \cite{health} and medicine \cite{medicine}.
 We distinguish between \textit{supervised learning}, in which an input-to-output relationship is described by a machine learning model that is trained from data, and \textit{unsupervised learning}, in which only inputs are considered, which are then e.g.~clustered. \cite{Diettreich,cluster} This work is concerned with supervised machine learning in the sense that we aim at replacing software components that would usually be realized by deterministic algorithms by machine learning models that are trained from data. Hence, we move from an \textit{algorithmic} to a \textit{data-driven approach}. 
        
        The analysis of massive volumes of raw data and measurements is often complicated, and can be much simplified with the help of \textit{scientific visualization}. {Scientific visualization} is a way of representing numerical data for its qualitative and quantitative analysis. 
        Scientific visualization follows the \textit{visualization pipeline}, which includes a sequence of steps  -- data enhancement, visualization mapping, and rendering -- to visualize data \cite{Hauser2018}. The visualization pipeline is further discussed in Section \ref{vis-pipeline}.
        Currently, scientific visualization is popular in different data domains like fluid data and astronomical data. \cite{Peikert2018}

        
       Traditional visualization tasks such as slicing, iso-contour extraction or velocity field visualization via glyphs and streamlines are nowadays easily realized by standard algorithmic software components. At the same time, highly problem-specific analysis tasks, such as vortex detection, turbulence classification and data-dependent segmentation is a much bigger challenge. Here, just having the domain knowledge is not enough for researchers. Instead, they are required to master the complexities of developing visualization algorithms to treat such tasks. This might be a hindrance for some researchers and could limit the extent of scientific discovery. 

Abstracting away the complexities of these algorithms into something more intuitive and simple would require the researchers only to deal with their well-known data and not with algorithmic details. Machine learning can be a potential way to achieve this goal. Machine learning models can be trained and optimized from (annotated) field-specific data. Integrating this into visualization would move the challenge of developing new visualization methods from the algorithmic side to an entirely data-driven side, thereby providing a pathway for many researchers to directly access the data visualization technology. \cite{K-LMa} At the same time, the scientific visualization community could profit from the regular tremendous improvements in machine learning technology.

Potential applications of coupling machine learning and visualization could be, as anticipated before, volume classification and data segmentation. \textit{Volume classification} defines a mapping from volume data values to color and brightness so that different regions can be identified with different colors \cite{K-LMa2}. Ma et al.\cite{K-LMa2} have demonstrated machine learning based high dimensional classification in a Magnetic Resonance Imaging(MRI) headset visualization. \textit{Segmentation} is concerned with classifying segments of region such that data behaviors of components within one segment are more similar to that segment's components than to other segments. 
        Researchers 
        are often interested in segmenting fluid simulations \cite{TILLMAN2012271} to study each region's properties in depth \cite{seg-CFD, seg-CFD2}. But fluid simulation data can be really complex, and it is not always simple to extract samples of relevance. For example, there may be different regions of interest within a single field and extracting meaningful data through simple segmentation may not be possible. Hence, machine learning based segmentation is a viable approach for these cases and can enable easier insights on each segment. 
        This work aims to explore the coupling of machine learning and scientific visualization with case studies using supervised segmentation and classification.


        The goal of this work is to achieve the coupling of machine learning into scientific visualization by extending the \textit{filters} in Paraview \cite{Paraview}, an open-source, multi-platform data analysis and visualization application, to also include filters with data-driven algorithms. Filters transform the input data according to the underlying algorithms. An example of a segmentation filter is illustrated in Figure~\ref{fig:bike-org-sc}.
        The extended filters, to be introduced in this work, load models \textit{pre-trained} with PyTorch \cite{Deepl}, a machine learning framework, and transform the data accordingly. \textit{Pre-trained} means that machine learning models have been already trained on some form of dataset and carry the resultant weights and biases (defined in Section~\ref{Background}). These models can then be loaded at any time and used for making predictions on datasets. This work  presents early use cases of coupling Paraview and PyTorch by introducing data-driven classification and segmentation filters for image and fluid data in Paraview. It further identifies future relevant use cases, where it may be beneficial to use such filters. 
        
        There has been a growing interest in the incorporation of machine learning into scientific visualization \cite{Bertini}. In Section~\ref{related-work}, relevant work introducing different variants of such combination are addressed. Some have integrated visualization in certain steps of machine learning to customize training of models  \cite{DeepTracker, TensorView, interactive-ml}, while some have integrated machine learning algorithms into visualization \cite{K-LMa3, PAVE, CobWeb, SlicerToolbox,Nvidia}. This work is concerned with the latter case by introducing data-driven classification and segmentation filters in Paraview, pretrained using PyTorch. To the best of the authors' knowledge, Paraview filters, till the time of writing, are based on deterministic algorithms to modify or extract slices of input data. This work extends the filters in Paraview to include algorithms based on data-driven transformations.  

         To summarize, the objective of this work is to implement plugins for data-driven filters in Paraview, 
         to assess the applicability of those filter plugins and to demonstrate that the coupling of Paraview and PyTorch is technically feasible. The rest of the document is structured as follows. Section~\ref{related-work} discusses relevant work in integrating machine learning and scientific visualization. Section~\ref{Background} gives a brief background of visualization in Paraview and machine learning concepts with respect to PyTorch. Section~\ref{sec:dataDrivenFilters} describes the requirements and implementation details of the plugins along with their use cases. The results are then presented in Section~\ref{Results}. Finally, Section~\ref{Conclusion} summarizes this work and points out its limitations and anticipated future work.
         



\section{Related work} \label{related-work}
     This section will discuss relevant work in the integration of scientific visualization and machine learning.
     
        As discussed in Section \ref{Introduction}, most works have focused on introducing machine learning into some areas of scientific visualization to enhance the visualization process and gain better insights on the given data. For example, Leventhal et al.\cite{PAVE} propose acceleration of path-tracing with neural networks, Chauhan et al.\cite{CobWeb} demonstrate a graphical user interface that uses machine learning based segmentation to enable region of interest selection, and Tzeng et al.\cite{K-LMa3} present a visualization system that learns to extract and track features in flow simulation.
        Similarly, machine learning extensions \cite{SlicerToolbox, Nvidia} for segmentation have been developed for \textit{3D Slicer} \cite{Slicer}, an open-source software application for medical image computing.


        Alternatively, there are also works that focus on using visualization in the training processes of machine learning algorithms. For example, Liu et al.\cite{DeepTracker} have introduced DeepTracker and Chen et al.\cite{TensorView} have presented TensorView as tools to visualize evolution of training models in Convolutional Neural Networks, and Li et al.\cite{interactive-ml} have presented a visualization approach to enable user interaction during training of models in machine learning. 
    
    
        Related work has introduced different variants of combination of machine learning and visualization, but there is not yet an integration of machine learning into the visualization pipeline of Paraview (e.g.~on fluid data) in terms of pre-trained filters. The\textit{ Supervised Segmentation Toolbox} \cite{SlicerToolbox} extension in \textit{3D Slicer} allows the user to train and use different classifiers like Support Vector Machines and Random Forests. The \textit{Nvidia AI-Assisted Annotation (AIAA)} \cite{Nvidia} plugin for \textit{3D Slicer} allows the user to perform segmentation of medical images based on pre-trained models. In the guided segmentation mode of AIAA, the user is required to give some input points for edges of region of interest to perform segmentation, and in the fully automatic segmentation mode of AIAA, the segmentation takes place without any user input. There is some potential overlap in two of the data-driven filter plugins of this work (introduced in Section \ref{use:img-seg} \& \ref{use:vel-seg}) and the aforementioned machine learning extensions in 3D Slicer. This is because both of these works incorporate machine learning based segmentation in visualization to enable better data analysis. However, this work is concerned with image as well as fluid data, and couples the interface of Paraview and PyTorch to visualize data driven transformations. To the extent of the author's knowledge, the current state-of-the-art in Paraview's filters are limited to transformations based on deterministic algorithms. This work extends these filters to include data driven algorithms, as outlined in Section \ref{Introduction}.


    
\section{Background} \label{Background}
  Paraview is based on the visualization library called \textit{The Visualization Toolkit} (VTK) \cite{vtkBook}. VTK serves as the backbone for data representation, processing and visualization in Paraview. This section gives an overview of \textit{data models} and the \textit{visualization pipeline} used in Paraview, and machine learning concepts (in context of PyTorch) relevant for this work.


  \subsection{Data models in VTK}\label{vtk-data-types}
  Datasets in VTK are represented by VTK \textit{data models}, formed of mesh and attributes. A mesh consists of cells and points. Points represent vertices of cells.
An attribute contains discrete values that can have any number of components. \cite{Paraview-data} The attributes with single component values are called \textit{scalars} and attributes with multiple components are called \textit{vectors}. Pressure and temperature in fluid flow models are examples of scalar data, and velocity is an example of vector data. Vector components are represented in VTK in the form of \textit{tuples}. These data attributes are stored as VTK \textit{data arrays}. This work deals with filters that operate on VTK data arrays having scalar and vector attributes. Paraview supports many VTK data models available \cite{Paraview-data}, but we here work with three of them: \textit{Image Data, Rectilinear Grid}, and \textit{Table}, explained below.
  
\textit{Image Data} is used to represent 1D lines, 2D images or 3D volumes. The \textit{dimension} is a 3-component vector $(nx, ny, nz)$ indicating number of points in x, y and z directions. The total number of points is $nx \cdot ny \cdot nz$, and all cells and points are spaced in regular intervals. 
 \cite{vtkBook}
%
\textit{Rectilinear Grid} represents a collection of points and cells arranged in a regular grid \cite{vtkBook}. The dimension, origin structure and numbering of cells and points is the same as Image Data, however the spacing between points in Rectilinear Grid may vary. 
%
\textit{Table} is a data structure for storing data in rows and columns. It requires every column to have the same number of entries. \cite{vtktable} 
  
  \begin{figure}[tb]
	\begin{center}
        \includegraphics*[scale=0.4]{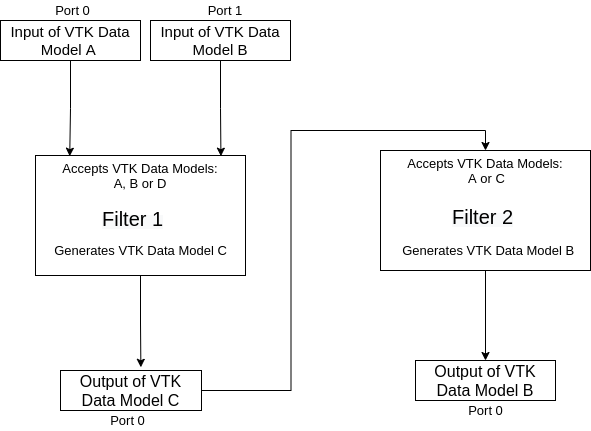}
	\end{center}
        \caption{Example of a visualization pipeline with two filters.}
         \label{vis-pipeline-fig}
\end{figure}  
    \subsection{Visualization pipeline} \label{vis-pipeline}
  The visualization pipeline is a way of representing the process of transforming information into visual data. It comprises of two components: data objects and process objects. \textit{Data objects} represent information and data. \textit{Process objects} operate on the data objects as input and generate output data objects. A \textit{filter} is a process object that transforms input data to generate output data. Filters consist of two parts: algorithm and executive objects. An \textit{algorithm object} processes information and data, and an \textit{executive object} decides when to execute the algorithm and what data should be processed. It requires one or more input data objects and produces one or more output data objects, depending on the algorithm used in the filter, as shown in Figure \ref{vis-pipeline-fig}. It also has \textit{parameters} whose values are supplied by the user, which influence the operation of the filter.
  
  The input data taken by the filter are retrieved from one or multiple \textit{input ports}, and the output data produced by the filter are stored in one or more \textit{output ports}. An input/output port refers to one logical input/output to the filter. For example, a filter generating a Red-Green-Blue (RGB) color image and corresponding \textit{binary mask} image can have two output ports, each port storing one image. It is also possible to connect a filter with multiple filters or other data objects. The pipeline can be designed with multiple data and process objects as long as the output of an \textit{upstream object} is compatible with the input of its \textit{downstream object}, for all objects connected in the pipeline \cite{vtkBook}. An \textit{upstream object} represents head of an arrow, and a \textit{downstream object} represents tail of an arrow. Data flows from an \textit{upstream object} to a \textit{downstream object}. An example of a visualization pipeline is shown in Figure \ref{vis-pipeline-fig}.

The components of the visualization pipeline mentioned above can be connected to create processing pipelines to perform tasks in Paraview, like in Figure \ref{vis-pipeline-fig}. The visual representation  of data is produced in the \textit{View} screen of Paraview. Paraview also allows changing \textit{color maps} while visualizing data on the screen. \textit{Color mapping} (also known as \textit{scalar mapping}) is a visualization technique which maps data to specific colors, and renders the image with those colors. A function used to map colors to the data is called \textit{transfer function} \cite{color-map}. Paraview provides an option to enable color mapping with the \textit{'Map Scalars'} option in the \textit{Properties Panel}. This option can be disabled if the user wishes to see Red-Green-Blue color view on the screen. The results of this work (see Section \ref{Results}) will include visualizations with \textit{scalar mapping}.

\subsection{Machine learning concepts}
We briefly review a few important notions from \textit{supervised learning}. In supervised learning, we are given a \textit{training set} of data $\{x_i,y_i\}_{i=1}^N\subset X\times Y$, where the $x_i$s are input samples and the $y_i$s are associated outputs or labels. The supervised machine learning task is to find a model or predictor $f_{{\alpha}}:X \rightarrow Y$, with a set of parameters ${\alpha}$, such that the \textit{loss} $L(f_{\alpha}(x_i),y_i)$, i.e.~a distance measure between the predicted output $f_{\alpha}(x_i)$ and the output $y_i$ from the training set, is minimized. Finding for a fixed model $f_{{\alpha}}$ its parameters is called \textit{training}. Evaluating the model is called \textit{prediction}. Ideally, a supervised machine learning model is able to provide predictions for unseen inputs, which ``best'' follow the relationship learned from the training data.

\begin{figure}[tb]
\begin{center}     
\includegraphics[scale=0.5]{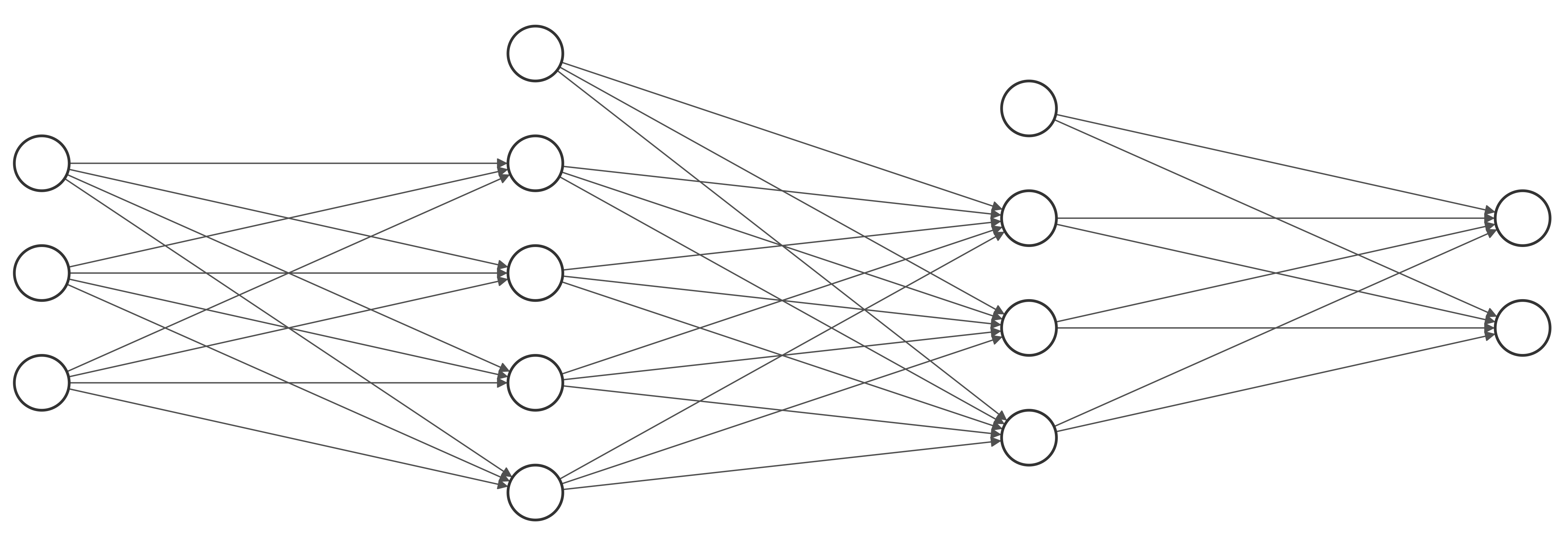}
\end{center}
    \caption{A neural network represented as a directed graph. }
    \label{fig:nn}
    \end{figure}

There are various supervised machine learning models. \textit{Artificial neural networks} (ANNs) are very popular. They are usually represented by directed graphs with nodes, where the nodes are called \textit{neurons}. The neurons are grouped in \textit{layers}, where the output of one layer is the input of the next layer \cite{Goodfellow2016}. Figure \ref{fig:nn} gives an example of such a graph, where we see on the left-hand side the \textit{input layer} and on the right-hand side the \textit{output layer}.

Indeed, the graph is just a visual representation for the underlying computational model that uses as model parameters so-called \textit{weights} and \textit{biases}. The model evaluation corresponds to propagating a given input $x$ through the different layers, i.e.~from left to right in Figure~\ref{fig:nn}. In each neuron, a weighted sum is computed over all connected neurons from the previous layer. Then, a predefined non-linear function known as \textit{activation function} is applied to the weighted sum. The training of an ANN is done using the \textit{backpropagation} algorithm, which is a combination of a \textit{gradient descent} -- type minimizer and an efficient way to calculate the gradients $\nabla L(f(x_i),y_i)$. Certainly, there is way more to say about general ANNs, however we leave it with this and refer the interested reader to standard literature in the field \cite{Hastie2009,Goodfellow2016,Geron2019}.

\textit{Convolutional Neural Networks (CNN)} are special types of ANNs, which have been initially introduced for image input data. In their traditional form, they consist of three types of layers: convolutional layers, pooling  layers  and fully connected layers.  The input layer holds pixel information of an image. The \textit{convolutional layers} can be understood as trainable filters and lead to a feature extraction for the input image. The \textit{pooling layer} downsamples the given input and fully-connected layers, such as shown in Figure~\ref{fig:nn}, will take the extracted features and do the main predictive task. \cite{Goodfellow2016} For both ANNs and CNNs, the definition of the exact structure of the different layers and their interaction is called \textit{network architecture}.

In this work, we will use simple ANNs and CNNs to carry out tasks of \textit{classification} and \textit{segmentation} on fluid fields such as velocity and pressure. Since the methodology of both tasks applied to fluid fields is similar of applying the same task to images, we briefly explain these tasks via their application to images. This mostly naturally extends to field data from fluid simulations.

\textit{Image classification} is the process of categorizing an image into one of several predefined classes or labels. An example of a \textit{binary} image classification is labelling an image whether it shows an animal or not. In classification via ANNs, each neuron in the output layer corresponds to one of the possible class labels. The value of the neuron is the \textit{confidence score} for the respective label. The output label with the highest confidence score is taken as predicted label of the classifier. For supervised image classification tasks, datasets with labelled data of thousands of categories have been developed like ImageNet \cite{imagenet} and LabelMe \cite{LabelMe}, along with CNN models like Alexnet \cite{Alexnet}, GoogLeNet \cite{Googlenet} and ResNet \cite{Resnet101}. 

\textit{Image segmentation} can be referred as image classification on pixel-level. It divides an image into regions based on image features like color, texture and shape \cite{ImageSeg}. All pixel information of an image is passed through the CNN model and it outputs a class label for each pixel. Then, each class is given a greyscale or RGB color value to visualize the segmented image. 
Datasets like PASCAL
VOC2012 \cite{pascal} and ImageNet \cite{imagenet} and models like FCN \cite{fcn-resnet101} and Deeplab \cite{deeplab} have been developed for performing supervised image segmentation. 

%

\subsection{Machine learning in PyTorch}\label{ml-pytorch}
The models for segmentation and classification to be presented in this work in Section \ref{Results} are trained using \textit{PyTorch}. {PyTorch} uses \textit{tensors} as its data structure, which are multi-dimensional matrices with all elements of single data type \cite{Deepl}, similar to \textit{NumPy} arrays. NumPy is a Python library for scientific computing that provides high performance multi-dimensional array objects and operations on the arrays \cite{numpy}. A NumPy array is a grid of values with all elements of a single data type. The training data used for building models in this work are in the form of NumPy arrays and are converted into tensors. PyTorch allows conversion between NumPy arrays and PyTorch tensors with external libraries \cite{Deepl}. 
%
    
The general workflow of training a model in PyTorch is as follows. The first step is to load training data as tensors. PyTorch provides several types of ANN layers, as well as activation and loss functions with the \texttt{torch.nn} module. After creating the network architecture and defining the loss function, the training data is fed to the neural network. Weights and biases are then trained using backpropagation using PyTorch's \texttt{torch.autograd} module. PyTorch also provides various optimized gradient-descent like minimizers, e.g.~the \textit{Adam algorithm} \cite{adam}, in \texttt{torch.optim} module. A pass of the entire training data set through the model one time is an \textit{epoch}. The dataset with input and expected output should be divided into validation and training dataset, such that the \textit{training dataset} is used to train the model, and the \textit{validation dataset} (separate from the training data) is used to validate the model during training. After training data for a number of epochs, the model can be used to predict output for training and validation dataset. This step is called \textit{model inference} \cite{pytorch-workflow}. The parameters of the model (weights and biases) can also be saved and then loaded any time for model inference.

  \section{Data-driven filters} \label{sec:dataDrivenFilters}
           This work intends to produce a visualization software component that demonstrates some use cases of data-driven filters in Paraview, to be discussed in Section \ref{usecases}. Paraview allows a software component called \textit{plugin} to extend its functionality or algorithmic capabilities. To enable aforementioned data-driven filters in Paraview for different use cases, Python-based filter plugins \cite{ParaviewPlugin} are developed in this work.  
In the following, we will first define a set of requirements that the anticipated plugins will need to fulfill. Then, we give a proposal of their design and algorithmic realization.
         
  \subsection{Requirements} \label{requirements}
\begin{figure}[tb]
\begin{center}
                \includegraphics*[scale=0.3]{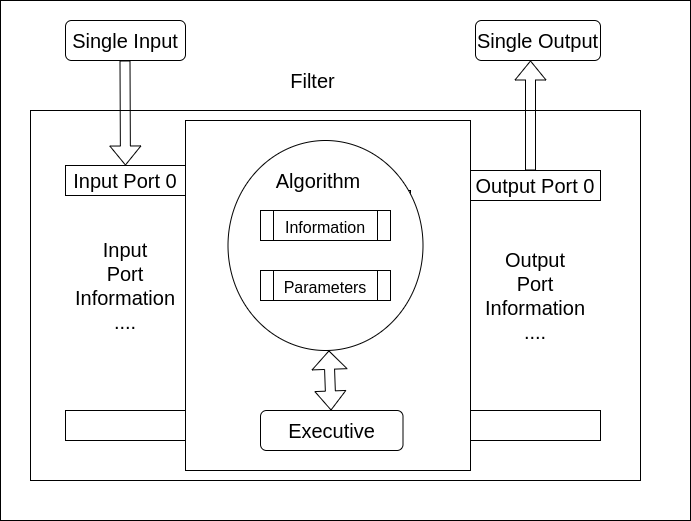}\quad\includegraphics*[scale=0.32]{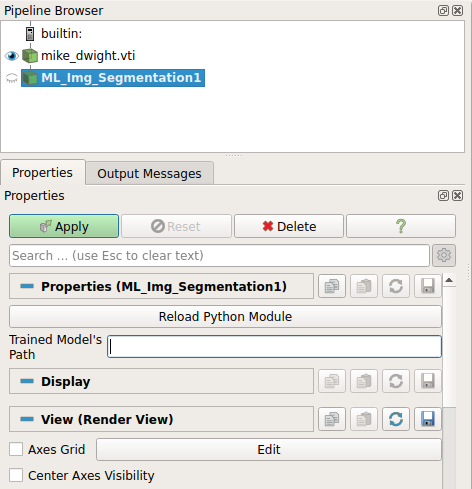}
\end{center}
                \caption{\textit{Left:} General filter architecture design of plugins introduced in this work. \textit{Right:} Screenshot of the Properties Panel the of the new image segmentation filter plugin.}
                 \label{design-filter-and-panel-fig}
            \end{figure}

\textbf{Plugin components.} The plugins provide two components: a \textit{filter} and its \textit{property panel} that extends the Paraview GUI and allows user interaction. The plugins are intended to work on a single port of input, resulting in output at a single port. The architecture design of the plugin to be used in this work is summarized in Figure \ref{design-filter-and-panel-fig} on the left-hand side. The plugins should, for now, support input of VTK data models: Image Data and Rectilinear Grid, and output of data models: Image Data, Rectilinear Grid and Table.  The plugins are designed corresponding to the Paraview plugin architecture \cite{ParaviewPlugin}, and follow the visualization pipeline discussed in Section \ref{vis-pipeline}.\newline
\textbf{Plugin functionality.} The plugins should allow the user to load the pre-trained models of their choice. This is achieved by enabling a text parameter in the \textit{Properties Panel} of Paraview, that requires the path to the pre-trained model. The plugins can handle both absolute path as well as path relative to Paraview's executable. The algorithm of the plugins then operate on the input data using the provided model. A screenshot of a plugin interface with the path parameter is provided in Figure~\ref{design-filter-and-panel-fig} on the right-hand side.\newline
\textbf{Compatibility of VTK data models in plugins.} According to the Paraview plugin mechanism \cite{ParaviewPlugin}, it is required to initially specify the VTK data models accepted for incoming input and outgoing output. However, the user may wish to apply a filter for a certain VTK data model that is different from the VTK data model the filter algorithm operates on.
    So, the idea of the plugins is that the core algorithm operates on only one VTK data model (which will be referred to as \textit{'base VTK data model'}) but is designed to accept multiple VTK data models as incoming input. The conversion from the other VTK data models to the base VTK data model is to be included internally in the plugin's algorithm. 
\newline 

\subsection{Generic filter design and implementation}
\begin{figure}[tb]
	\begin{center}
                \includegraphics*[scale=0.3]{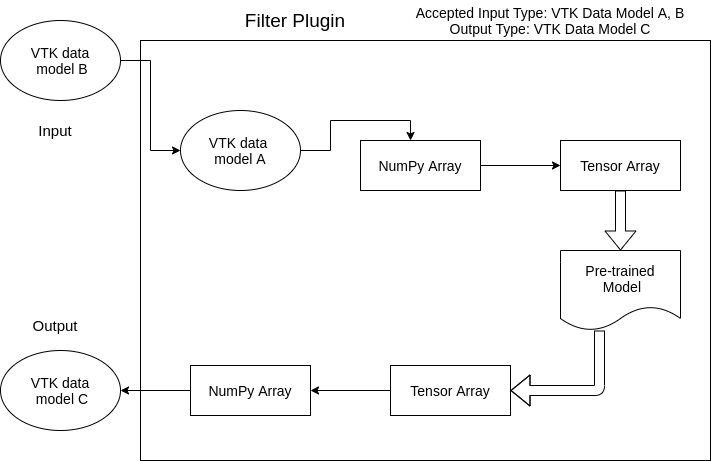}
	\end{center}
                \caption{VTK and PyTorch use different data representations. This requires to include appropriate data conversion algorithms in the general structure of all anticipated plugins. 
                \\
 }
                 \label{fig:data-conversion}
            \end{figure}

The general structure of the new data-driven filter plugins is given in Figure~\ref{fig:data-conversion}. Here, the major challenge is to appropriately map between the various data representations in VTK/Paraview and PyTorch.

The plugins are designed to feed data attributes of the VTK data models as input to pre-trained models. These data attributes are stored as VTK data arrays, however the pre-trained PyTorch model works with Tensor arrays as outlined in Section \ref{ml-pytorch}. So, a conversion between VTK data arrays and Tensor arrays is also required to couple the interface between Paraview and PyTorch. For this conversion, a NumPy interface is used in between due to the existing libraries offering conversion support between NumPy and VTK data arrays as well as NumPy and Tensor arrays. Thus, the conversion of Tensor to NumPy to VTK data arrays and vice-versa is performed with the modules: \texttt{'torch.from\_numpy'}, \texttt{'paraview.vtk.util.numpy\_support'} and\linebreak \texttt{'vtkmodules.numpy\_interface.dataset\_adapter'}.
    
    The plugins use type checking in different steps to ensure a proper connection is established between different interfaces. 
    This flow of data conversion is demonstrated with the generic plugin structure in Figure \ref{fig:data-conversion}. Here, the accepted input VTK data models are A and B, where A is the base VTK data model, and the plugin supports conversion from data model B to A.
    
In Algorithm~\ref{alg:generic}, we summarize the core functionality of the new filters for various input and output types, which will be further described in the next section. It is easy to observe that the previously described requirements on the plugin functionality and on the compatibility of the VTK data models are fulfilled.

\begin{algorithm}[tb]
	    \SetAlgoLined
            \SetKwData{Left}{left}
            \SetKwData{This}{this}
            \SetKwData{Up}{up}
            \SetKwFunction{Union}{Union}
            \SetKwFunction{FindCompress}{FindCompress}
            \SetKwInOut{Input}{Input}
            \SetKwInOut{Output}{Output}
            \SetKwBlock{Ad}{Requesting input data:}{}
            \SetKwBlock{Vn}{Data conversion:}{}
            \SetKwBlock{Pr}{Model Inference:}{}
            \SetKwBlock{Pre}{Pre-processing data:}{}
            \SetKwBlock{Ou}{Generating output data:}{}
            \Input{VTK Image Data / VTK Rectilinear Grid}
            \Output{VTK Image Data / VTK Rectilinear Grid / VTK Table }
            \Ad{
             Get the path to pre-trained model from the user \;
	     Potentially get the path to further auxiliary data from the user \;
             }
             \Vn{
             Extract the pixel/point values (VTK data array) from the input \;
                  Convert the VTK data array to NumPy array \;
            }
             \Pr{
             Load the pre-trained model \;
             \Pre{ 
             Resize the NumPy array to the size require by the model \;
             Convert the NumPy array into a Tensor array \;
             Potentially normalize the Tensor array appropriately \;
             }
             Feed the pre-processed array to the model \;  
             Convert the predicted output (Tensor) to a NumPy array \;
             Map the ML model's output to the appropriate information (color values / confidence percentage / \ldots) \;
        }
             \Vn{
             Convert the mapped NumPy array to a VTK data array \;}
             \Ou{
         Assign the VTK data array as Point / Row Data for output \;
            }
        
            \caption{\label{alg:generic}Generic filter plugin algorithm}
            
            \end{algorithm}

Paraview plugins can be distributed in binary shared libraries or as a Python file. The Python-based plugins for this work have been developed with Paraview 5.8.1, PyTorch 1.6.0, NumPy 1.19.1 and Python 3.7.4. 
 
     \subsection{Use cases} \label{usecases}
At total of four plugins have been designed to show potential use cases of the Paraview-Pytorch coupling.  All these plugins follow the same pipeline design mentioned before in Section \ref{requirements}, but the accepted input/output data models and the core algorithm of the filter varies. Two of the plugins provide filters for the task of segmentation and two of the plugins provide filters for the task of classification. For each task, we first have a prototypic filter, where we use image data as input and a pre-trained segmentation/classification model that is downloaded from PyTorch. While image data is certainly not the future intended input for a data-driven filter in Paraview, we use this test case as a sanity check and to shed light on the possibility to download and use pre-trained models in the context of Paraview. Similarly, for each task, we have as a second use case a fluid visualization oriented filter, which takes fluid data fields as input and provides a simplistic segmentation/classification example via a machine learning model trained by the authors (see Section~\ref{Results}). Overall, these use cases aim to show only simple prototypes of segmentation and classification within Paraview. In the future, more versatile machine learning models for e.g.~turbulence classification or vortex region detection need to be developed. These can then be simply loaded into the already existing plugins. 

         \subsubsection{Image segmentation}
         \label{use:img-seg}
         This use case intends to show how supervised image segmentation can be carried out within Paraview. It further emphasizes on the possibility of using existing pre-trained models in Paraview, which might be particularly interesting for users who don't want to train their own models and want to apply existing models on their datasets. The plugin accepts only VTK Image Data as input, and has one parameter for the path to the trained model. It is designed to work for models that take 3-channel images (RGB color images) of size $256 \times 256$, which are normalized with the ImageNet mean and standard deviation(STD) where, the mean is  $(0.485, 0.456, 0.406)$ and the standard deviation is $(0.229, 0.224, 0.225)$ \cite{imagenet}. In case of a greyscale image (1 channel), the plugin converts it to RGB image by considering the greyscale intensity for all three channels. For example, a greyscale pixel value $5$ is converted to a RGB pixel value $(5,5,5)$. So, the plugin's algorithm uses this pre-processing step to prepare the data for model inference. The pre-processing of input data is carried out using the \texttt{'torch.transforms'} module. The model classifies each pixel of pre-processed image into one of the pre-defined classes and each pixel is represented by the corresponding color value defined for that class. The details of the machine learning model's architecture will be discussed in Section \ref{result:img-seg}.

         \subsubsection{Fluid velocity field segmentation} \label{use:vel-seg}
        This use case demonstrates that the concept of segmentation is not limited to images and can also be extended to fluid data. The velocity information of fluid simulations is stored in Paraview as a tuple in each point of the Rectilinear Grid. The velocity values for all the points serve as input to the trained model. The machine learning model, which is further discussed in Section~\ref{res:vel-seg}, classifies each point into one of the pre-defined classes. The model takes as input a \textit{flattened} array of size $2500$ with each element having a 3 component tuple. This plugin works for VTK Rectilinear Grid. It has two user parameters: path to the trained model, and path to the class definition (architecture) of the model. 
        Here, the reader should be reminded that the to be discussed model is only a simplified version of a segmentation, and intends to motivate users to build their own models similarly for their datasets. Due to the modularity of the developed plugin, this will be easily possible.

         \subsubsection{Image classification}
This use case intends to show the possibility of using pre-existing models for classification within the context of Paraview. The plugin for this use case works for both VTK Image Data and VTK Rectilinear Grid, with Rectilinear Grid as its \textit{base VTK data model}. It takes two user parameters: the path to trained model, and the path to file containing names of class labels. The class labels in the file should be in the same order in which the training labels were enumerated. This plugin is designed to work for models that take 3-channel images (RGB color images) of size $256 \times 256$, normalized with ImageNet mean and standard deviation mentioned in Section \ref{use:img-seg}. The model classifies the pre-processed image into one of the pre-defined labels. The details of the model's architecture will be discussed in Section \ref{res:img-cls}. This plugin also supports greyscale to RGB image conversion as mentioned in Section \ref{use:img-seg}.

         \subsubsection{Pressure field classification}
         Similar to Section \ref{use:vel-seg}, the concept of classification is extended to fluid data stored in the form of rectilinear grids. A scalar value is stored at each point in the grid, which represents pressure at that point. The pressure values for all the points serve as  input to the model. The model is designed to take a flattened input array of size $2500$ with single component elements. The model architecture and other training details will be discussed in Section \ref{res:pressure-classify}. This plugin works for VTK Rectilinear Grid, and has two user parameters: the path to trained model, and the path to class definition of the model. The model trained for this use case will only be a simple version of \textit{binary} classification, and can be extended to more advanced classifiers by users.

  \section{Results} \label{Results}
  This section discusses the results of using the aforementioned plugins in Paraview. It is to be reminded that image segmentation and classification may not be the perfect applications for scientific visualization and the plugins for these use cases mainly aim to highlight the possibility of using existing trained models within Paraview. So, the models used in this work for image segmentation and classification are downloaded from PyTorch. On the other hand, the models for velocity segmentation and pressure classification are trained by the authors. The data used for training these models is extracted from several two-dimensional flow simulations of a lid-driven cavity for changing boundary velocity and Reynolds number. The fluid solver that produces the data implements a discretization of the Navier-Stokes equations based on \cite{Navier-stokes}. It is to be noted that the models constructed by the authors are "synthetic" models, only intended for demonstrating the viability of using self-developed models within Paraview. 
  
  \subsection{Image segmentation} \label{result:img-seg}
      This work uses DeepLab-v3 \cite{deeplab} as pre-trained model for image segmentation.
      The Deeplab-v3 model is a deep learning model built by Google with a ResNet-101 \cite{fcn-resnet101} backbone. It is trained on a subset of the COCO train2017 dataset with 21 labels and is available in a pre-trained form in PyTorch. \cite{fcn-resnet101, deeplab}
      
      The used model requires the input of 3-channel (RBG) images normalized with the ImageNet \cite{imagenet} mean and standard deviation. The images need to be pre-processed to be in the right format for inference using the model. A forward pass through the model's network results in 21 channels of output, and for each pixel, the index of the maximum output weight indicates its predicted class. The predicted class indices are then mapped to color values such that each class is represented by a different color.  
    	
\begin{figure}[t]
\begin{center}
            \includegraphics*[width=0.45\linewidth]{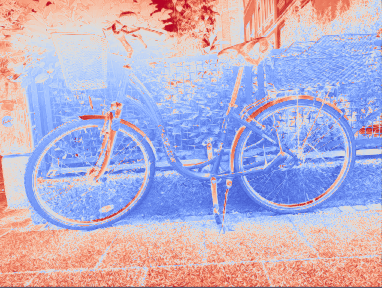}
            \includegraphics*[width=0.45\linewidth]{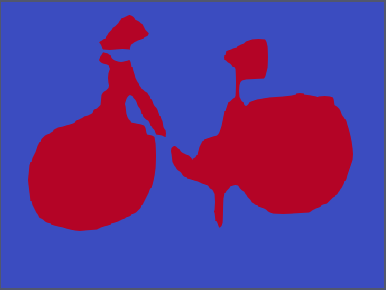}
\end{center}
            \caption{\label{fig:bike-org-sc}Result of the image segmentation plugin in Paraview with scalar mapping. \textit{Left:} Input image that goes into the newly developed filter. \textit{Right:} The output segmentation provided by the data-driven filter.}
    \end{figure}

      One exemplary result of the developed image segmentation plugin is presented in Figure \ref{fig:bike-org-sc}. The figure displays on the left-hand side the visualization view of an input image in Paraview with applied scalar mapping (defined in Section \ref{Background}). On the right-hand side in the same figure, the output of the plugin using the DeepLab-v3 model is given. Obviously, the segmentation detects individually the bicycle and the background.
    	
    
\subsection{Velocity segmentation} \label{res:vel-seg}
	    The architecture used to train the model for velocity segmentation is a simple neural network that contains 3 hidden linear layers with $80$, $40$ and $10$ neurons respectively and uses \textit{tanh} as activation function. The input layer contains 3 neurons and the output layer contains 2 neurons. The training input is an array of size $2500$ with 3-component vector(velocity) values of a $50 \times 50$ grid. The training output labels a point as 1 if the velocity magnitude in that point is greater than a specified value, referred to as  \textit{threshold value}, otherwise the point is labelled as 0. This classification is done for every point so the resulting grid gives a segmented grid, similar to a segmented image in Section \ref{result:img-seg}. Obviously, the proposed segmentation is very similar to a data-driven replacement of the standard threshold filter provided by Paraview, hence it truly oversimplifies a segmentation for a velocity field. A more dedicated model would take the full velocity field as input and would e.g.~characterize different parts of the field to be turbulent or laminar. 
	    
	    Our simplistic model was trained by the authors using the aforementioned neural network architecture with a \textit{threshold value} $0.01$. The model was trained using the \textit{Cross Entropy Loss} \cite{Goodfellow2016} and the \textit{Adam optimization algorithm} \cite{adam} with a learning rate of $5\cdot 10^{-4}$. The dataset was divided such that $80\%$ of the dataset was used as training data and 20\% as validation data, and the model was trained for 5000 epochs. The velocity data for training was taken from the mentioned lid-driven cavity flow simulation scene with Reynold's number $10$, and strength of the inflow $2.0$ for $50 \times 50$ grid. This data was generated from a simulation model based on Navier-Stokes equations \cite{Navier-stokes}, as outlined before. This simulation scene represents a highly \textit{viscous fluid}.
	    
\begin{figure}[tb]
	\begin{center}
            \includegraphics*[width=0.31\linewidth]{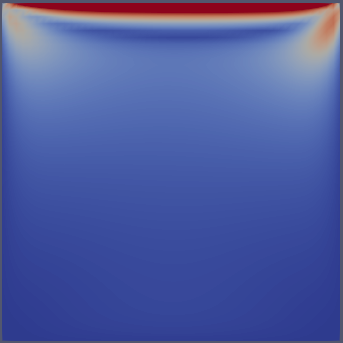}
	    \includegraphics*[width=0.31\linewidth]{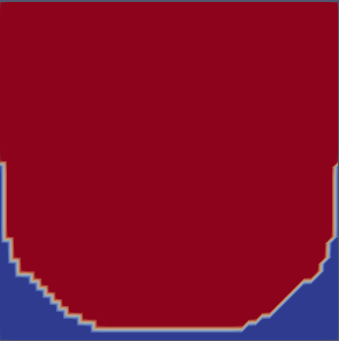}
	    \includegraphics*[width=0.31\linewidth]{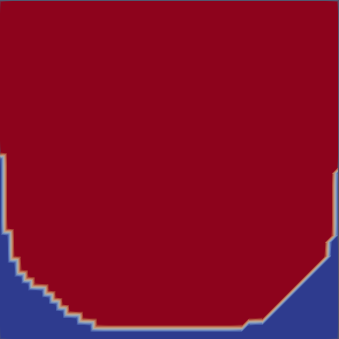}
	\end{center}
            \caption{Simplistic velocity field segmentation (velocity magnitude thresholding with threshold $0.01$) for time step $t=20$ with visualizations of the input \textit{(left)}, the ground truth \textit{(middle)} and the data-driven segmentation \textit{(right)}.}
            \label{fig:vel-t10}
    \end{figure}

        An exemplary result of the Paraview filter plugin developed for velocity segmentation is presented in Figure \ref{fig:vel-t10}, where the first one considers the velocity field for time step $t=0$ and the second one considers it for time step $t=20$. On the left-hand side of both figures, the input flow field is given. In the middle, the exact threshold segmentation is given, which we approximate by the segmentation model. On the right-hand side, the predicted threshold segmentation by our very simplistic model is provided.

\subsection{Image classification} \label{res:img-cls}
      The pre-trained image classification model \textit{Alexnet} \cite{Alexnet} is used for this application. Alexnet is a CNN architecture with 5 convolutional layers and 3 fully connected layers. Alexnet is provided with PyTorch pre-trained on the \textit{ImageNet} \cite{imagenet} dataset with 1000 labels.
     
     A forward pass through the model's network results in 1000 channels of output, and the class with the maximum weight among the 1000 channels indicates its predicted class. The resulting weights are mapped into confidence scores of the model. The predicted classes and the confidence scores of the highest 10 scores are displayed in \textit{Spreadsheet View} as a table in Paraview. 
        
\begin{figure}[tb]
	\begin{center}
            \includegraphics*[width=0.327\linewidth]{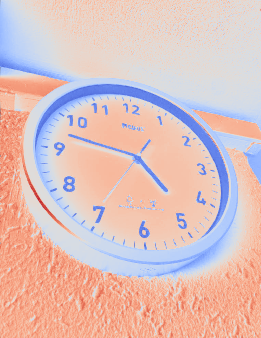}
            \includegraphics*[width=0.49\linewidth]{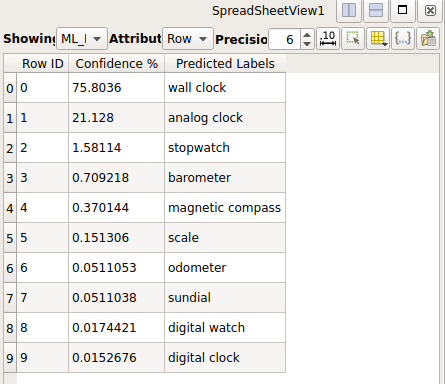}
	\end{center}
            \caption{Exemplary result of the newly implemented image classification plugin in Paraview. The input image of a wall clock with applied scalar mapping \textit{(left)} is correctly classified as wall clock in the Spreadsheet View \textit{(right)}.}
            \label{fig:strawberry-org}
    \end{figure}

     An exemplary result of the plugin developed for image classification is presented in Figure \ref{fig:strawberry-org}, with an input image visualized by Paraview with color mapping on the left-hand side and the \textit{Spreadsheet View} of the output of the newly developed filter on the right-hand side.
        
\subsection{Pressure classification}
\label{res:pressure-classify}

	    The architecture used to train the model for pressure classification is a neural network that contains 2 hidden linear layers with $50$ and $20$ neurons and uses \textit{tanh} as activation function. The input layer contains $2500$ neurons and the output layer contains 2 neurons.  The training input is a flattened array of size $2500$ with scalar pressure values of a grid of size $50 \times 50$. The training output is 1 if the maximum pressure value of the grid is greater than the threshold value, otherwise the output is 0. The weights of the two output neurons of the model are then transformed to be between the range of 0 and 1 using the \textit{Softmax} function \cite{Goodfellow2016}. The index of the neuron with the highest transformed value indicates the predicted class. If the resulting index is 0, the class is labelled as 'Low' and if the resulting index is 1, the class is labelled as 'High'. Certainly, this is again only a very simplistic classifier just designed for demonstration purposes.
	    
	    A classification model was constructed by the authors using the aforementioned neural network architecture with a threshold value of $5.0$. The model was again trained over $500$ epochs using the Cross Entropy Loss and the Adam optimization algorithm with a learning rate of $5\cdot 10^{-4}$. $80\%$ of the underlying dataset was used as training data and 20\% as validation data. The employed training data was again taken from the lid-driven cavity flow simulation scene with Reynold's number $1000$, and inflow velocity $-1.0$ on a $50 \times 50$ grid. 
	    
\begin{figure}[h]
	\begin{center}
            \includegraphics*[width=0.382\linewidth]{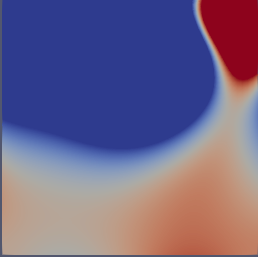}
            \includegraphics*[width=0.5\linewidth]{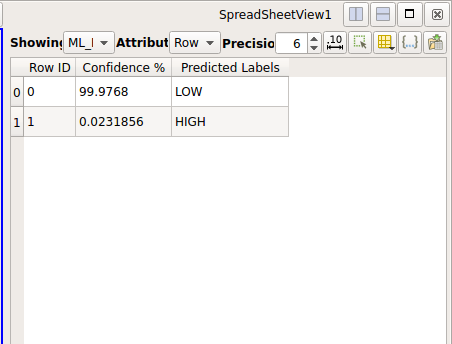}
	\end{center}
          \caption{Result of the pressure classification plugin using the model trained with threshold value $5.0$ in Paraview. \textit{Left:} Input pressure visualization with scalar mapping. \textit{Right:} The plugin's output in Spreadsheet View.}
            \label{fig:p-R1000-T5}
    \end{figure}

        An exemplary result of the plugin developed for pressure classification is presented in Figure \ref{fig:p-R1000-T5}.
It shows the input pressure field at time $t=5$ on the left hand-side and the plugin's output in Spreadsheet View on the right-hand side.
        

  \section{Conclusion \& future work} \label{Conclusion}
    Four filter plugins were introduced in this work to showcase the proof of concept of coupling Paraview with PyTorch in order to profit from data-driven filters in visualization. The usability of the conceptual idea was demonstrated through image segmentation, velocity segmentation, image classification and pressure classification respectively. Image segmentation and image classification were considered as sanity check and proof of principle for the use of pre-trained machine learning models in Paraview, noting that they may not be the ultimate goals in scientific visualization. Moreover, prototypes for plugins that allow the introduction of data-driven segmentation and classification filters for fluid data like pressure and velocity were proposed. As a first step towards truly research-oriented data-driven filters, two simplistic machine learning models for velocity segmentation and pressure classification were developed. The developed plugins are available as an open-source project \cite{githubProject}, which invites researchers to develop and load more sophisticated machine learning models. 
As a future application, the analysis of e.g.~turbulence in fluid flows could be considered \cite{Navier-stokes}. So, the models used in this work could be replaced by neural networks models that can e.g.~predict Reynold's number or classify turbulence \cite{turbulence}. 

On a technological side, the plugins presented by this work deal with simple data sets on a single port of input and output, and are mostly used for demonstrating the viability of using pre-trained machine learning models within Paraview. As designing neural networks and creating relevant datasets has not been the scope of this work, this work was mainly focused on showcasing the possibilities of extending the algorithmic capabilities of Paraview towards data-driven filters. Visualizing scientific data may require the use of advanced features like parallelization, multiple ports, animation etc.~present in Paraview. These features have not been utilized by this work, and could be part of potential future work.












\bibliographystyle{abbrv}

\bibliography{scivis}
\end{document}